%% file: main.tex
\documentclass[11pt,a4paper]{article}
\usepackage[hyperref]{emnlp-ijcnlp-2019}
\usepackage{times}
\usepackage{latexsym}

\usepackage{url}

\aclfinalcopy %

\usepackage{graphicx}
\usepackage{amsmath}

\DeclareMathOperator*{\argmin}{argmin}
\usepackage{bm, upgreek}
\usepackage{amssymb}
\usepackage{array, booktabs}
\usepackage{tabularx}
\usepackage{multirow}
\usepackage{mathtools}
\usepackage{arydshln}
\usepackage{tabularx}
\usepackage{diagbox}

\usepackage{subcaption}

\usepackage{soul}
\usepackage{threeparttable}

\newif\ifcomments
\ifcomments
     
     \providecommand{\ms}[1]{{\protect\color{cyan}{\bf [Minjoon: #1]}}}
     \providecommand{\jm}[1]{{\protect\color{orange}{\bf [Jaemin: #1]}}}

\else
     \providecommand{\ms}[1]{}
     \providecommand{\jm}[1]{}
\fi

\usepackage[ruled,vlined,linesnumbered]{algorithm2e}

\SetCommentSty{mycommfont}

\usepackage{multicol}
\usepackage{footnote}
\makesavenoteenv{tabular}
\makesavenoteenv{table}

\usepackage{makecell}

\input{00-commands}

\title{
    Mixture Content Selection for Diverse Sequence Generation
}

\author{
    Jaemin Cho$^1$\thanks{\ \  Most work done during internship at Clova AI.}
    \qquad
    Minjoon Seo$^{2,3}$
    \qquad
    Hannaneh Hajishirzi$^{1,3}$ \\
    Allen Institute for AI$^1$
    \quad
    Clova AI, NAVER$^2$
    \quad
    University of Washington$^3$ \\
    { \tt heythisischo@gmail.com \quad \{minjoon,hannaneh\}@cs.washington.edu }
}

\date{}

\begin{document}
\maketitle
\begin{abstract}
\input{00-absract}

\end{abstract}

\section{Introduction}

\input{01-intro}

\section{Related Work} \label{sec:related}

\input{02-related}

\section{Method} \label{sec:overview}

\input{03-method-overview}

\input{03-method-training}

\section{Experimental Setup} \label{sec:exp}
\input{04-exp}

\section{Results} \label{sec:results}

\input{05-results}

\section{Conclusion}
\input{06-con}

\section*{Acknowledgment}
This research was supported by Allen Distinguished Investigator Award, the Office of Naval Research under the MURI grant N00014-18-1-2670, Samsung GRO, and gifts from Allen Institute for AI and Google. We also thank the members of Clova AI,
UW NLP, and the anonymous reviewers for their
insightful comments.

\bibliography{main}
\bibliographystyle{acl_natbib}

\end{document}

\appendix

\section{Attention visualization on Abstract Summarization}

Appendices ({\em i.e.} supplementary material in the form of proofs, tables,
or pseudo-code) should be {\bf uploaded as supplementary material} when submitting the paper for review.
Upon acceptance, the appendices come after the references, as shown here. Use
\verb|\appendix| before any appendix section to switch the section
numbering over to letters.

%% file: 00-commands.tex
\newcommand{\Ours}{\textsc{Selector}}

\newcommand{\eb}{\mathbf{e}}

\newcommand{\hb}{\mathbf{h}}

\newcommand{\mb}{\mathbf{m}}

\newcommand{\yb}{\mathbf{y}}

\newcommand{\xb}{\mathbf{x}}

%% file: 00-absract.tex
Generating  diverse  sequences is  important in many NLP applications such as question generation or summarization that exhibit semantically one-to-many relationships between  source and the target sequences. %
We present a method to explicitly separate diversification  from generation using a general plug-and-play module (called \Ours) that wraps around and guides an existing encoder-decoder model. The diversification stage  uses  a mixture of experts to sample different binary masks on the source sequence for diverse content selection. The generation stage uses a standard encoder-decoder model given each selected content from the source sequence.
Due to the non-differentiable nature of discrete sampling and the lack of ground truth labels for binary mask, we leverage  a  proxy for ground-truth mask and adopt stochastic hard-EM for training. 
In question generation (SQuAD) and abstractive summarization (CNN-DM), our method demonstrates significant improvements in accuracy, diversity and training efficiency, including state-of-the-art top-1 accuracy in both datasets, 6\% gain in top-5 accuracy, and 3.7 times faster training over a state-of-the-art model.
Our code is publicly available at \url{https://github.com/clovaai/FocusSeq2Seq}.

%% file: 01-intro.tex
Generating target sequences given a source sequence has applications in a wide range of problems in NLP with different types of relationships between the source and target sequences. 
For instance, paraphrasing or machine translation exhibit a one-to-one relationship because the source and the target should carry the same meaning. %
On the other hand, summarization or question generation exhibit one-to-many relationships because a single source often results in diverse target sequences with different semantics. Fig.~\ref{fig:example} shows different questions that can be generated from a given passage.  

\input{01-intro-f1}

Encoder-decoder models~\cite{Cho2014} are widely used for sequence generation, most notably in machine translation where neural models are now often almost as good as human translators in some language pairs.
However, a standard encoder-decoder often shows a poor performance when it attempts to produce multiple, diverse outputs. Most recent methods for diverse sequence generation leverage  diversifying decoding steps through alternative search algorithms~\cite{Fan2018,Vijayakumar2018} or mixture of decoders~\cite{He2018, Shen2019}. These methods promote diversity at the decoding step,  while a more focused selection of the source sequence can lead to diversifying the semantics of the generated target sequences.

In this paper, we present a method for diverse generation that separates diversification and generation stages.
The diversification stage leverages content selection to map the source to multiple sequences, where each mapping is modeled by \textit{focusing} on different tokens in the source (one-to-many mapping).
The generation stage uses a standard encoder-decoder model to generate a target sequence given each selected content from the source (one-to-one mapping).
We present a generic module called \Ours\ that is specialized for diversification. This module can be used as a plug-and-play to an arbitrary encoder-decoder model for generation without architecture change.

The \Ours\ module leverages a mixture of experts \cite{Jacobs1991, Eigen2014} to identify diverse key contents to focus on during generation. Each mixture samples a sequential latent variable modeled as a binary mask on every source sequence token.
Then an encoder-decoder model generates multiple target sequences given these binary masks along with the original source tokens.
Due to the non-differentiable nature of discrete sampling, we adopt stochastic hard-EM for training \Ours. To mitigate the lack of ground truth annotation for the mask (content selection), we use the overlap between the source and target sequences as a simple proxy for the ground-truth mask.

We experiment on question generation and abstractive summarization tasks
and show that our method achieves the best trade-off between accuracy and diversity over previous models on SQuAD \cite{Rajpurkar2016} and CNN-DM \cite{Hermann2015, Nallapati2016, See2017} datasets.
In particular, compared to the recently-introduced mixture decoder \cite{Shen2019} that also aims to diversify outputs by creating multiple decoders, our modular method not only demonstrates better accuracy and diversity, but also trains 3.7 times faster.

\input{01-intro-f2.tex}

%% file: 01-intro-f1.tex
\begin{figure}
\begin{center}
\begin{small}
\begin{tabularx}{\linewidth}{p{\linewidth-1em}}
\toprule

\textbf{Source Passage}: in december 1878 , tesla \ul{left graz and severed all relations with his family} to hide the fact that he dropped out of school . \\
\textbf{Target}: what did tesla do in december 1878? \\
\midrule
\textcolor{blue}{\textbf{Focus 1}}:
in december 1878 , \textcolor{blue}{tesla} \ul{left graz and severed all relations with his family} to hide the fact that he dropped out of school. \\
(Ours) $\Rightarrow$ what did tesla do? \\
\midrule

\textcolor{Green}{\textbf{Focus 2}}:
in \textcolor{Green}{december 1878 , tesla} \ul{left graz and severed all relations with his family} to hide the fact that he dropped out of school. \\
(Ours) $\Rightarrow$ what did tesla do in december 1878? \\
\midrule

\textcolor{red}{\textbf{Focus 3}}:
in december 1878 , \textcolor{red}{tesla} \ul{left graz and severed all relations with his family} to \textcolor{red}{hide} the fact that he \textcolor{red}{dropped} out of \textcolor{red}{school} . \\
(Ours) $\Rightarrow$ what did tesla do to hide he dropped out of school? \\

\bottomrule
\end{tabularx}
\end{small}
\end{center}
\caption{
    Sample questions produced by our method from given passage-answer pair (answer is underlined). Our method generates diverse questions, by selecting different tokens to focus (colored) in contrast to  3-mixture decoder \cite{Shen2019} that generates 3 identical questions: ``what did tesla do to hide the fact that he dropped out of school?''.
}
\label{fig:example}
\end{figure}

%% file: 01-intro-f2.tex
\begin{figure*}[!h]
\includegraphics[width=\textwidth]{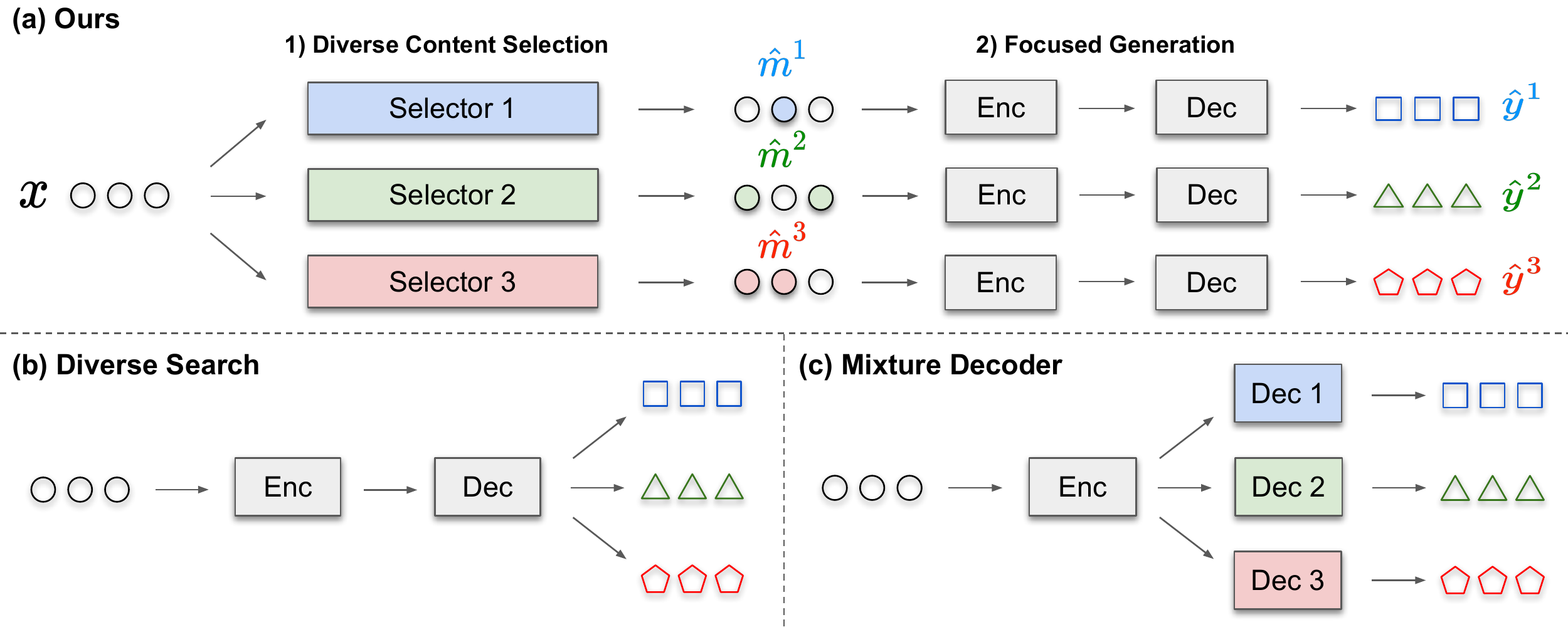}
\caption{
    Overview of diverse sequence-to-sequence generation methods.
    (a) refers to our two-stage approach described throughout Section. \ref{sec:overview},
    (b) refers to search-based methods \cite{Vijayakumar2018, Li2016, Fan2018},
    and (c) refers to mixture decoders \cite{Shen2019, He2018}.
}
\label{figure:methods_horizontal}
\end{figure*}

%% file: 02-related.tex
\paragraph{Diverse Search Algorithms}
Beam search, the most commonly used search algorithm for decoding, is known to produce samples that are short, contain repetitive phrases, and share majority of their tokens.
Hence several methods are introduced to  diversify search algorithms for decoding.
\citet{Graves2013, Chorowski2016} tune temperature hyperparameter in softmax function.
\citet{Vijayakumar2018, Li2016} penalize similar samples during beam search in order to obtain diverse set of samples.
\citet{Cho2016} adds random noise to RNN decoder hidden states.
\citet{Fan2018} sample tokens from top-k tokens at each decoding step.
Our method is orthogonal to these search-based strategies,
in that they diversify
\textit{decoding}
while our method diversifies
which content to be focused 
during \textit{encoding}.
Moreover, our empirical results show diversification with stochastic sampling hurts accuracy significantly.

\paragraph{Deep Mixture of Experts}
Several methods adopt a deep mixture of experts (MoE)~\cite{Jacobs1991, Eigen2014} to diversify decoding steps.
\citet{Yang2018} introduce soft mixture of softmax on top of the output layer of RNN language model.
\citet{He2018, Shen2019} introduce mixture of decoders with uniform mixing coefficient to improve diversity in machine translation. %
Among these, the closest to ours is the mixture decoder~\cite{Shen2019} that also adopts hard-EM for training, where a minimum-loss predictor is assigned to each data point, which is also known as multiple choice learning~\cite{Guzman-Rivera2012, Lee2016}.
While \citet{Shen2019} makes RNN decoder as a MoE, we make \Ours\ as a MoE to diversify content selection and let the encoder-decoder models one-to-one generation.
As shown in our empirical results, our method achieves a better accuracy-diversity trade-off while reducing training time significantly.

\paragraph{Variational Autoencoders}
Variational Autoencoders (VAE) \cite{Kingma2013} are used 
for diverse generation in several tasks, such as language modeling \cite{Bowman2016},
machine translation \cite{Zhang2016, Su2018, Deng2018, Shankar2019},
and conversation modeling \cite{Serban2017, Wen2017, Zhao2017, Park2018, Wen2018, Gu2019}.
These methods sample diverse latent variables from an approximate posterior distribution, but often suffer from a posterior collapse where the sampled latent variables are ignored \cite{Bowman2016, Park2018, Kim2018, Dieng2018, Xu2018, He2019, Razavi2019}. 
This is also observed in our initial experiments and \citet{Shen2019}, where MoE-based methods significantly outperforms VAE-based methods because of the posterior collapse. %
Moreover, we observe that sampling mixtures  makes training more stable compared to stochastic sampling of latent variables. 
Furthermore, our latent structure as a sequence of binary variables  is different from most VAEs which use a fixed-size continuous latent variable.
This gives a finer-grained control and interpretability on where to focus, especially when source sequence is long.

\paragraph{Diversity-Promoting Regularization}
Adding regularization to objective functions is used to diversify generation. 
\citet{Li2016a} introduce a term maximizing mutual information between source and target sentences.
\citet{Chorowski2016, Kalyan2018} introduce terms enforcing knowledge transfer among similar annotations. 
Our work is orthogonal to these methods and can potentially benefit from adding these regularization terms to our objective function.

\paragraph{Content Selection in NLP}
Selecting important parts of the context has been an important step in NLP applications \cite{Reiter2000}.
Most recently, \citet{Ke2018, Min2018} conduct soft-/ hard-selection of key parts from source passages for question answering.
\citet{Zhou2017a} use soft gating on the source document encoder for abstractive summarization.
\citet{Li2018a} guide abstractive summarization models with off-the-shelf keyword extractors.
The most relevant work to ours are \citet{Subramanian2018} and \citet{Gehrmann2018}.
\citet{Subramanian2018} use a pointer network \cite{Vinyals2015a} for extracting key phrases for question generation and \citet{Gehrmann2018} use content selector to limit copying probability for abstractive summarization. 
The main purpose of these approaches is to enhance accuracy, while our method uses diverse content selection to enhance both accuracy and diversity (refer to our empirical results).
Additionally, our method allows models to learn how to utilize information from the selected content, whereas  \citet{Gehrmann2018}  manually limit the copying mechanism on non-selected contents.

%% file: 03-method-overview.tex
In this paper, we focus on sequence generation tasks such as question generation and summarization that have one-to-many relationship.
More formally, given a source sequence $\xb = (x_1 \ldots x_S) \in \mathcal{X}$, our goal is to model a conditional multimodal distribution for the target sequence $p(\yb|\xb)$ that assigns
high values on $p(\yb^1|\xb) \ldots\ p(\yb^K|\xb)$ for $K$ valid mappings $\xb \rightarrow \yb^1 \ldots\ \xb \rightarrow \yb^K$.
For instance, Fig. \ref{fig:example} illustrates $K=3$ different valid questions generated for a given passage.
 
For learning a multimodal distribution, it is not appropriate to use a generic encoder-decoder~\cite{Cho2014} that minimizes for the expected value of the log probability of all the valid mappings. This can lead to a suboptimal mapping that is in the middle of the targets but not near any of them. %
As a solution, we propose to (1) introduce a latent variable called \emph{focus} that factorizes the distribution into two stages, \emph{select} and \emph{generate} (Section~\ref{subsec:focus}), and (2) independently train the factorized distributions (Section~\ref{subsec:training}).

\subsection{Select and Generate}
\label{subsec:focus}
In order to factorize the multimodal distribution into the two stages (\emph{select} and \emph{generate}),
we introduce a latent variable called \emph{focus}.
The intuition is that in the \emph{select} stage we sample several meaningful \emph{focus}, each of which indicates which part of the source sequence should be considered important.
Then in the \emph{generate} stage, each sampled focus biases the generation process towards being conditioned on the focused content.

Formally, we model \emph{focus} with a sequence of binary variable, each of which corresponds to each token in the input sequence, i.e. $\mb$ = $\{m_1 \ldots m_S\} \in \{0,1\}^{S}$.
The intuition is that $m_t=1$ indicates $t$-th source token $x_t$ should be \textit{focused} during sequence generation.
For instance, in Fig. \ref{fig:example}, colored tokens (green, red, or blue) show that different tokens are focused (i.e. values are 1) for different focus samples (out of 3).
We first use the latent variable $\mb$ to factorize 
 $p(\yb|\xb)$,
\begin{equation}
\label{eq:marginalization}
    p(\yb|\xb) =
    \mathbb{E}_{\mb \sim p_{\phi}(\mb |\xb)}
    [p_{\theta}(\yb|\mb,\xb)]
\end{equation}
where
$p_{\phi}(\mb|\xb)$ is \emph{selector} and 
$p_{\theta}(\yb|\xb, \mb)$ is \emph{generator}.
The factorization separates focus selection from generation so that
modeling multimodality (diverse outputs) can be solely handled in the \textit{select} stage and
\textit{generate} stage can solely concentrate on the generation task itself.
We now describe each component in more details.

\paragraph{Selector} 
In order to directly control the diversity of the \Ours's outputs, we model it as a hard mixture of experts (hard-MoE) \cite{Jacobs1991, Eigen2014}, where each expert specializes in focusing on different parts of the source sequences.
In Fig. \ref{fig:example} and \ref{figure:methods_horizontal},
focus produced by each \Ours\ expert is colored differently.
We introduce a multinomial latent variable $z \in \mathbb{Z}$, where $\mathbb{Z} = \{1 \ldots K\}$,
and let each focus $\mb$ be assigned to one of K experts with uniform prior $p(z|\xb) = \frac{1}{K}$.
With this mixture setting, $p(\mb|\xb)$ is recovered as follows.
\begin{equation}
\label{eq:marignalization_full_mixture}
\begin{split}
    p(\mb|\xb) & = \mathbb{E}_{z \sim p(z |\xb)} [ p_{\phi}(\mb|\xb, z) ] \\
            & = \frac{1}{K} \sum^{1 \ldots K}_{z}  p_{\phi}(\mb|\xb, z)
\end{split}
\end{equation}

We model \Ours\ with a single-layer Bi-directional Gated Recurrent Unit (Bi-GRU) \cite{Cho2014} followed by two fully-connected layers and a Sigmoid activation.
We feed (current hidden state $\hb_t$, first and last hidden state $\hb_1, \hb_S$, and expert embedding $\eb_z$) to a fully-connected layers (FC).
Expert embedding $\eb_z$ is unique for each expert and is trained from a random initialization.
From our initial experiments, we found this parallel focus inference to be more effective than auto-regressive pointing mechanism \cite{Vinyals2015a, Subramanian2018}.
The distribution of the focus conditioned on the input $\xb$ and the expert id $z$ is the Bernoulli distribution of the resulting values,
\begin{equation}
\label{eq:selector}
    \begin{split}
        (\hb_1 \ldots \hb_S) & = \text{Bi-GRU}(\xb) \\
        o_{t}^{z} & = \sigma(\text{FC}([\hb_t; \hb_1; \hb_S; \eb_{z}])) \\
         p_{\phi}(m_t |\xb, z) & = \text{Bernoulli}(o_{t}^{z})
    \end{split}
\end{equation}
 
To prevent low quality experts from \emph{dying} during training, we let experts share all parameters, except for the individual expert embedding $\eb_z$.
We also reuse the word embedding of the generator in the word embedding of \Ours\ to promote cooperative knowledge sharing between \Ours\ and generator.
With these parameter sharing techniques, adding a mixture of \Ours\ experts increases only a slight amount of parameters (GRU, FC, and $\eb_z$) to sequence-to-sequence models.

\paragraph{Generator} For maximum diversity, we sample one focus from each \Ours\ expert to approximate $ p_{\phi}(\mb|\xb)$.
For a deterministic behavior, we threshold $o_{t}^{z}$ with a hyperparameter $th$ instead of sampling from the Bernoulli distribution.
This gives us a list of focus $\mb^1 \dots \mb^K$ coming from $K$ experts.
Each focus $\mb^z = (m_1^z \ldots m_S^z)$ is encoded as embeddings and concatenated with the input embeddings of the source sequence $\xb = (x_1 \ldots x_S)$. 
An off-the-shelf generation function such as encoder-decoder can be used for modeling $p(\yb|\mb, \xb)$, as long as it accepts a stream of input embeddings.
We use an identical generation function with $K$ different focus samples to produce $K$ different diverse outputs.

%% file: 03-method-training.tex
\input{03-method-a1}

\subsection{Training}\label{subsec:training}
Marginalizing the Bernoulli distribution in Eq. \ref{eq:selector} by enumerating all possible focus is intractable since the cardinality of focus space $2^S$ grows exponentially with source sequence length $S$.
Policy gradient~\cite{Williams1992, Yu2017} or Gumbel-softmax~\cite{Jang2017, Maddison2017} are often used to propagate gradients through a stochastic process, but we empirically found that these do not work well.
We instead create \emph{focus guide} and use it to independently and directly train the \Ours\ and the generator.
Formally, a focus guide $\mb^\text{guide} = (m^\text{guide}_1 \ldots m^\text{guide}_S)$ is
a simple proxy of whether a source token is \textit{focused} during generation.
We set $t$-th focus guide $m^\text{guide}_t$ to $1$ if $t$-th source token $x_t$ is \textit{focused} in target sequence $\yb$ and $0$ otherwise.
During training, $\mb^\text{guide}$ acts as a target for \Ours\ and is a given input for generator (teacher forcing). During inference, $\hat{\mb}$ is sampled from \Ours\ and fed to the generator.

In question generation, we set $m^\text{guide}_t$ to $1$ if there is a target question token which shares the same word stem with passage token $x_t$. Then we set $m^\text{guide}_t$ to $0$ if $x_t$ is a stop word or is inside the answer phrase. %
In abstractive summarization, we generate focus guide  using copy target generation %
by~\citet{Gehrmann2018}, where they set source document token $x_t$ is \textit{copied} if it is part of the longest possible subsequence that overlaps with the target summary.

Alg. \ref{alg:training} describes the overall training process, which first uses stochastic hard-EM \cite{Neal1998, Lee2016} for training the \Ours\ and then the canonical MLE for training the generator.

\paragraph{E-step}
(line 2-5 in Alg. \ref{alg:training})
we sample focus from all experts and compute their losses $-\log p_{\phi}(\mb^\text{guide}|\xb, z)$.
Then we choose an expert $z^\text{best}$ with minimum loss.
\begin{equation}
z^\text{best} = \argmin\limits_z -\log p_{\phi}(\mb^\text{guide}|\xb, z)
\end{equation}

\paragraph{M-step}
(line 6 in Alg. \ref{alg:training})
we only update the parameters of the chosen expert $z^\text{best}$.

\paragraph{Training Generator}
(line 7-8 in Alg. \ref{alg:training})
The generator is independently trained using conventional teacher forcing, which minimizes $-\log p_{\theta}(\yb|\xb,\mb^\text{guide})$.

%% file: 03-method-a1.tex
\SetEndCharOfAlgoLine{}
\begin{algorithm}
\caption{Training\\
(N: Dataset size, K: Number of mixtures)}
\label{alg:training}
    \SetAlgoLined
    \KwData{$\mathcal{D} = \{ (\xb^{(i)}, \yb^{(i)}, \mb^\text{guide\ (i)}) \}^{N}_{i=1}$}
    \For{$i \in \{1 \ldots N \}$}{
        \tcc{Selector $p_{\phi}(\mb|\xb,z)$\ E\text{-}step}
        \For{$z \in \{1 \ldots K \}$}{
            $L^{(i)\ z}_{\text{select}}$ = $-\log p_{\phi}(\mb^{\text{guide}\ (i)}|\xb^{(i)}, z)$\;
        }
        $z^{\text{best}\ (i)} = \argmin\limits_z L^{(i)\ z}_{\text{select}}$\;
        \tcc{Selector $p_{\phi}(\mb|\xb,z)$\ M\text{-}step}
        $\phi_{z^{\text{best}\ (i)}} = \phi_{z^{\text{best}\ (i)}} - \alpha \nabla_{\phi_{z^{\text{best}\ (i)}}} L^{(i)\ z^{\text{best}\ (i)}}_{{\text{select}}}$\;
        \tcc{Generator $p_{\theta}(\yb|\xb,\mb)$ Update}
        ${L^{(i)}}_{\text{gen}}$ = $-\log p_{\theta}(\yb^{(i)}|\xb^{(i)}, \mb^\text{guide\ (i)})$\;
        $\theta = \theta - \alpha \nabla_{\theta} L^{(i)}_{\text{gen}}$
    }
\end{algorithm}

%% file: 04-exp.tex
We describe our experimental setup for question generation (Section~\ref{subsec:QG}) and abstractive summarization (Section~\ref{subsec:SM}).
For both tasks, we use an off-the-shelf, task-specific encoder-decoder-based model for the generator and show how adding \Ours\ can help to diversify the output of an arbitrary generator. To evaluate the contribution of \Ours\, we additionally compare our method with previous diversity-promoting methods as the baseline (Section~\ref{subsec:diversity}).

\subsection{Question Generation}
\label{subsec:QG}
Question generation is the task of generating a question from a passage-answer pair. Answers are given as a span in the passage (See Fig.\ref{fig:example}).

\paragraph{Dataset} We conduct experiments on SQuAD \cite{Rajpurkar2016} and use the same dataset split of \citet{Zhou2017}, resulting in  86,635, 8,965, and 8,964 source-target pairs for training, validation, and test, respectively.
Both source passages and target questions are single sentences.
The average length of source passage and target question are 32 and 11 tokens.

\paragraph{Generator}
We use NQG++ \cite{Zhou2017} as the generator, which is an RNN-based encoder-decoder architecture with copying mechanism \cite{Gulcehre2016}.

\subsection{Abstractive Summarization}
\label{subsec:SM}
Abstractive summarization is the task of generating a summary sentence from a source document that consists of multiple sentences. 

\paragraph{Dataset}
We conduct experiments on the non-anonymized version of CNN-DM dataset~\cite{Hermann2015, Nallapati2016, See2017},
whose training, validation, test splits have size of 287,113, 13,368, and 11,490 source-target pairs, respectively. 
The average length of the source documents and target summaries are 386 and 55 tokens. 
Following \citet{See2017}, we truncate source and target sentences to 400 and 100 tokens during training.

\paragraph{Generator}
We use Pointer Generator (PG)~\cite{See2017} as the generator for summarization, which also leverages RNN-based encoder-decoder architecture and copying mechanism, and uses coverage loss to avoid repetitive phrases.

\subsection{Baselines}\label{subsec:diversity}
For each task, we compare our method with other techniques which promote diversity at the decoding step. In particular, we compare with recent diverse search algorithms including
Truncated Sampling \cite{Fan2018},
Diverse Beam Search \cite{Vijayakumar2018},
and Mixture Decoder \cite{Shen2019}.
We implement these methods with NQG++ and PG.
For each method, we generate K = (3 and 5) hypotheses from each source sequence.
For search-based baselines \cite{Fan2018, Vijayakumar2018}, we select the top-k candidates after generation. 
For mixture models (\citet{Shen2019} and ours), we conduct greedy decoding from each mixture for fair comparison with search-based methods in terms of speed/memory usage.

\vspace{.1cm}
\noindent{\bf Beam Search}
This baseline keeps $K$ hypotheses with highest log-probability scores at each decoding step.

\vspace{.1cm}
\noindent{\bf Diverse Beam Search}
This baseline adds a diversity promoting term to log-probability when scoring hypotheses in beam search,
Following \citet{Vijayakumar2018}, we use hamming diversity and diversity strength $\lambda = 0.5$.

\vspace{.1cm}
\noindent{\bf Truncated Sampling}
This baseline randomly samples words from top-10 candidates of the distribution at the decoding step \cite{Fan2018}.

\vspace{.1cm}
\noindent{\bf Mixture Decoder}
This baseline constructs a hard-MoE of K decoders with uniform mixing coefficient (referred as hMup in \citet{Shen2019}) and conducts parallel greedy decoding.
All decoders share all parameters but use different embeddings for start-of-sequence token.

\vspace{.1cm}
\noindent{\bf Mixture Selector (Ours)}
We construct a hard-MoE of K \Ours{}s with uniform mixing coefficient that infers K different focus from source sequence.
Guided by K focus, generator conducts parallel greedy decoding.

\subsection{Metrics: Accuracy and Diversity}
\label{sec:metrics}
We use metrics introduced by previous works \cite{Ott2018, Vijayakumar2018, Zhu2018} to evaluate the  diversity promoting approaches. These metrics are extensions over  BLEU-4 \cite{Papineni2002} and ROUGE-2 $F_1$-score \cite{Lin2004} and aim to evaluate the trade-off between accuracy and diversity.

\paragraph{Top-1 metric ($\Uparrow$)}
This measures the Top-1 accuracy among the generated K-best hypotheses. The accuracy is measured using a corpus-level metric, i.e., BLEU-4 or ROUGE-2. 

\paragraph{Oracle metric ($\Uparrow$)}
This measures the quality of the target distribution coverage among the Top-K generated target sequences \cite{Ott2018, Vijayakumar2018}.
Given an optimal ranking method (oracle), this metric measures the upper bound of Top-1 accuracy by comparing the best hypothesis with the target. %
Concretely, we %
generate hypotheses $\{\hat{\yb}^1 \ldots \hat{\yb}^K\}$ from each source $\xb$ and
keep the hypothesis $\hat{\yb}^{best}$ that achieves the best sentence-level metric with the target $\yb$.
Then we calculate a corpus-level metric with the greedily-selected hypotheses $\{\hat{\yb}^{(i),best}\}^N_{i=1}$ and references $\{\yb^{(i)}\}^N_{i=1}$.

\paragraph{Pairwise metric ($\Downarrow$)}
Referred as self-~\cite{Zhu2018} or pairwise-~\cite{Ott2018} metric, this measures the within-distribution similarity.
This metric computes the average of sentence-level metrics between all pairwise combinations of hypotheses $\{\hat{\yb}^1 \ldots \hat{\yb}^K\}$ generated from each source sequence $\xb$.
Low pairwise metric indicates high diversity between generated hypotheses.

\subsection{Human Evaluation Setup}
We ask Amazon Mechanical Turkers (AMT) to compare our method with the baselines. For each method, we generate three questions / summaries from 100 passages sampled from SQuAD / CNN-DM test set.
For every pair of methods, annotators are instructed to pick a set of questions / summaries that are more {\it diverse}.
To evaluate {\it accuracy}, they see one question / summary selected out of 3 questions / summaries with highest log-probability from each method. They are instructed to select a question / summary that is more coherent with the source passage / document. Each annotator is asked to choose either a better method (resulting in ``win'' or ``lose'') or ``tie'' if their quality is indistinguishable.
Diversity and accuracy evaluations are conducted separately, and every pair of methods are presented to 10 annotators\footnote{We conducted 12 human evaluations in total: 2 tasks x 3 baselines x 2 criteria. See Table \ref{table:human_qg} and \ref{table:human_sm}.}.

\subsection{Implementation details}
\label{exp_detail}
For all experiments, we tie the weights \cite{Press2017} of the encoder embedding, the decoder embedding, and the decoder output layers.
This significantly reduces the number of parameters and training time until convergence.
We train up to 20 epochs and select the checkpoint with the best oracle metric.
We use Adam \cite{Kingma2015} optimizer with learning rate 0.001 and momentum parmeters $\beta_1 = 0.9$ and $\beta_2 = 0.999$.
Minibatch size is 64 and 32 for question generation and abstractive summarization.
All models are implemented in PyTorch \cite{Paszke2017} and trained on single Tesla P40 GPU,
based on NAVER Smart Machine Learning (NSML) platform \cite{Kim2018a}. 
\paragraph{Question Generation}
Following \citet{Zhou2017},
we use 256-dim hidden states for each direction of Bi-GRU encoder,
512-dim hidden states for GRU decoder,
300-dim word embedding initialized from GloVe~\cite{Pennington2014},
vocabulary of 20,000 most frequent words,
16-dim embeddings for three linguistic features (POS, NER and word case) respectively.
\paragraph{Abstractive Summarization}
Following \citet{See2017},
we use 256-dim hidden states for each direction of Bi-LSTM encoder and LSTM decoder,
128-dim word embedding trained from scratch,
and vocabulary of 50,000 most frequent words.
Following \citet{See2017}, we train our model to generate concatenation of target summaries and split it with periods.
\paragraph{\Ours}
The GRU has the same size as the generator encoder,
and the dimension of expert embedding $\eb_z$ is 300 for NQG++, and 128 for PG.
From simple grid search over [0.1, 0.5], we obtain focus binarization threshold $th$ 0.15. 
The size of focus embedding for the generator is 16.

%% file: 05-results.tex
\paragraph{Diversity vs. Accuracy Trade-off}
Tables~\ref{table:qg} and \ref{table:sm} compare our method with different diversity-promoting techniques in question generation and abstractive summarization. The tables show that our mixture \Ours\ method outperforms all baselines in Top-1 and oracle metrics and achieves the best trade-off between diversity and accuracy. Moreover, both mixture models are superior to all search-based methods in the trade-off between diversity and accuracy. 
Standard and diverse beam search methods  score low both in terms of accuracy and diversity.  Truncated sampling shows the lowest self-similarity (high diversity), but it achieves the lowest score on Top-1 accuracy. %
Notably, our method scores state-of-the-art BLEU-4 in question generation on SQuAD and ROUGE comparable  to state-of-the-art methods in abstractive summarization in CNN-DM (See also Table \ref{table:single_expert} for state-of-the-art results in CNN-DM).

\input{05-results-t-qg.tex}

\input{05-results-t-sm.tex}

\paragraph{Diversity vs. Number of Mixtures}
Here we compare the effect of number of  mixtures in our \Ours\ and Mixture Decoder~\cite{Shen2019}.  Tables \ref{table:qg} and \ref{table:sm} show that pairwise similarity increases (diversity $\Downarrow$) when the number of mixtures increases for Mixture Decoder. While we observe a similar trend for \Ours\ in the question generation task, Table~\ref{table:sm} shows the opposite effect in the summarization task i.e., pairwise similarity decreases (diversity $\Uparrow$) for \Ours. 

The abstractive summarization task on CNN-DM has a target distribution with more modalities than question generation task on SQuAD, which is more difficult to model.
We speculate that our \Ours\ improves accuracy by focusing on more modes of the output distribution (diversity $\Uparrow$), whereas Mixture Decoder tries to improve the accuracy by concentrating on fewer modalities of the output distribution (diversity $\Downarrow$).
    
\paragraph{Upper Bound Performance}
The bottom rows of Tables~\ref{table:qg} and~\ref{table:sm} show the upper bound performance of \Ours\ by feeding focus guide to generator during test time. In particular, we assume that the mask is the ground truth overlap between input and target sequences at test time.  
The gap between the oracle metric (top-k accuracy) and the upper bound is very small for question generation.  This indicates that the top-k masks  for question generation include the ground truth mask. Future work involves improving the content selection stage for the summarization task.  %

\input{05-results-attn_fig.tex}

\input{05-results-t-human.tex}

\paragraph{Human Evaluation} 
Table~\ref{table:human_qg} and \ref{table:human_sm} show the human evaluation in two tasks of questions generation and summarization, comparing sequences generated by \Ours\ with the diversity-promoting baselines: Diverse Beam, Truncated Sampling and Mixture Decoder.
The table shows that our method significantly outperforms all three baselines in terms of both diversity and accuracy with statistical significance.

\paragraph{Comparison with State-of-the-art}

\label{para:single-expert}
Table~\ref{table:single_expert} compares the performance of \Ours\ with the state-of-the-art  bottom-up content selection of~\citet{Gehrmann2018} in abstractive summarization. %
 \Ours\ passes focus embeddings at the decoding step, whereas the bottom-up selection method only uses the masked words for the copy mechanism. %
We set $K$, the number of mixtures of \Ours{}, to $1$ to directly compare it with the previous work (Bottom-Up~\cite{Gehrmann2018}). 
We observe that \Ours\ not only outperforms Bottom-Up in every metric, but also achieves a new state-of-the-art ROUGE-1 and ROUGE-L on CNN-DM. Moreover, our method scores state-of-the-art BLEU-4 in  question  generation  on  SQuAD (Table~\ref{table:qg}).

\input{05-results-t-single-expert.tex}

\paragraph{Efficient Training}
\label{effecient_training}
Table \ref{table:training_time} shows that \Ours\ trains up to 3.7 times faster than mixture decoder \cite{Shen2019}.
Training time of mixture decoder linearly increases with the number of decoders, while parallel focus inference of \Ours\ makes additional training time negligible.

\input{05-results-t-training-time.tex}

\paragraph{Qualitative Analysis}

To analyze how the generator uses the selected content, we visualize attention heatmap of NQG++ in Fig.\ref{fig:attention} for question generation.
The figure shows different attention mechanisms depending on three different focuses inferred by different \Ours\ experts.

%% file: 05-results-t-qg.tex
\begin{table}[t]
\resizebox{\columnwidth}{!}{%
    \begin{threeparttable}
        \begin{tabular}{l||c|c|c}
            \hline
            \multicolumn{1}{c||}{\multirow{2}{*}{Method}}
                & BLEU-4 & Oracle  & Pairwise  \\
                & (Top-1 ) & (Top-K) & (Self-sim) \\
            \hline
            NQG++ & 13.27 & - & -\\ %
            \hline
            \multicolumn{4}{c}{Search-based Methods} \\
            \hline
            3-Beam & 13.590 & 16.848 & 67.277 \\
            5-Beam & 13.526 & 18.809 & 74.674 \\
            3-D. Beam & 13.696 & 16.989 & 68.018 \\
            5-D. Beam & 13.379 & 18.298 & 74.795 \\
            3-T. Sampling & 11.890 & 15.447 & \textbf{37.372} \\
            5-T. Sampling & 11.530 & 17.651 & 45.990 \\
            \hline
            \multicolumn{4}{c}{Mixture of Experts + Greedy Decoding} \\
            \hline
            3-M. Decoder & 14.720 & 19.324 & 51.360 \\
            5-M. Decoder & 15.166 & 21.965 & 58.727 \\
            3-M. \Ours\ (Ours) & \textbf{15.874} & 20.437 & 47.493 \\
            5-M. \Ours\ (Ours) & 15.672 & \textbf{22.451} & 59.815 \\
            \hline
            \multicolumn{4}{c}{Focus Guide during Test Time} \\
            \hline
            5-Beam + Focus Guide & \multicolumn{2}{c|}{24.580} & - \\
            \hline
        \end{tabular}
    \end{threeparttable}}
\caption{
    {\bf Question generation results: }Comparison of diverse generation methods on SQuAD.
    The score of NQG++ (top row) is from \citet{Zhou2017},
    and the rest are from our experiments using NQG++ as a generator.
    Method prefixes are the numbers of generated questions for each passage-answer pair.
    Best scores are bolded.
}
\label{table:qg}
\end{table}

%% file: 05-results-t-sm.tex
\begin{table}[t]
    \resizebox{\columnwidth}{!}{%
    \begin{tabular}{l||c|c|c}
        \hline
        \multicolumn{1}{c||}{\multirow{2}{*}{Method}}
            & ROUGE-2 & Oracle & Pairwise \\
            & (Top-1) & (Top-K) & (Self-sim) \\ %
        \hline
        PG & 17.28 & - & - \\
        \hline
        \multicolumn{4}{c}{Search-based Methods} \\
        \hline
        3-Beam & 16.533 & 18.509 & 85.598 \\
        5-Beam & 16.634 & 19.442 & 84.765 \\
        3-D. Beam & 16.667 & 18.722 & 85.496 \\
        5-D. Beam & 16.632 & 19.659 & 84.043 \\
        3-T. Sampling & 12.914 & 17.068 & 17.306 \\
        5-T. Sampling & 13.049 & 19.161 & \textbf{16.720} \\
        \hline
        \multicolumn{4}{c}{Mixture of Experts + Greedy Decoding} \\
        \hline
        3-M. Decoder & 15.854 & 21.214 & 43.168 \\
        5-M. Decoder & 16.104 & 21.801 & 67.196 \\
        3-M. \Ours\ (Ours) & 17.930 & 21.316 & 51.092 \\
        5-M. \Ours\ (Ours) & \textbf{18.309} & \textbf{22.511} & 47.280 \\
        \hline
        \multicolumn{4}{c}{Focus Guide during Test Time} \\
        \hline
        5-Beam + Focus Guide & \multicolumn{2}{c|}{42.757} & - \\
        \hline
    \end{tabular}
    }
    \caption{
        {\bf Summarization results: }Comparison of diverse generation methods on CNN-DM.
        The score of PG (top row) is from \citet{See2017},
        and the rest are from our experiments using PG as a generator.
        Method prefixes are the numbers of generated summaries for each document.
        Best scores are bolded.
        }
    \label{table:sm}
\end{table}

%% file: 05-results-attn_fig.tex
\begin{figure*}[t]\begin{center}
\includegraphics[
                width=\textwidth,
                 ]{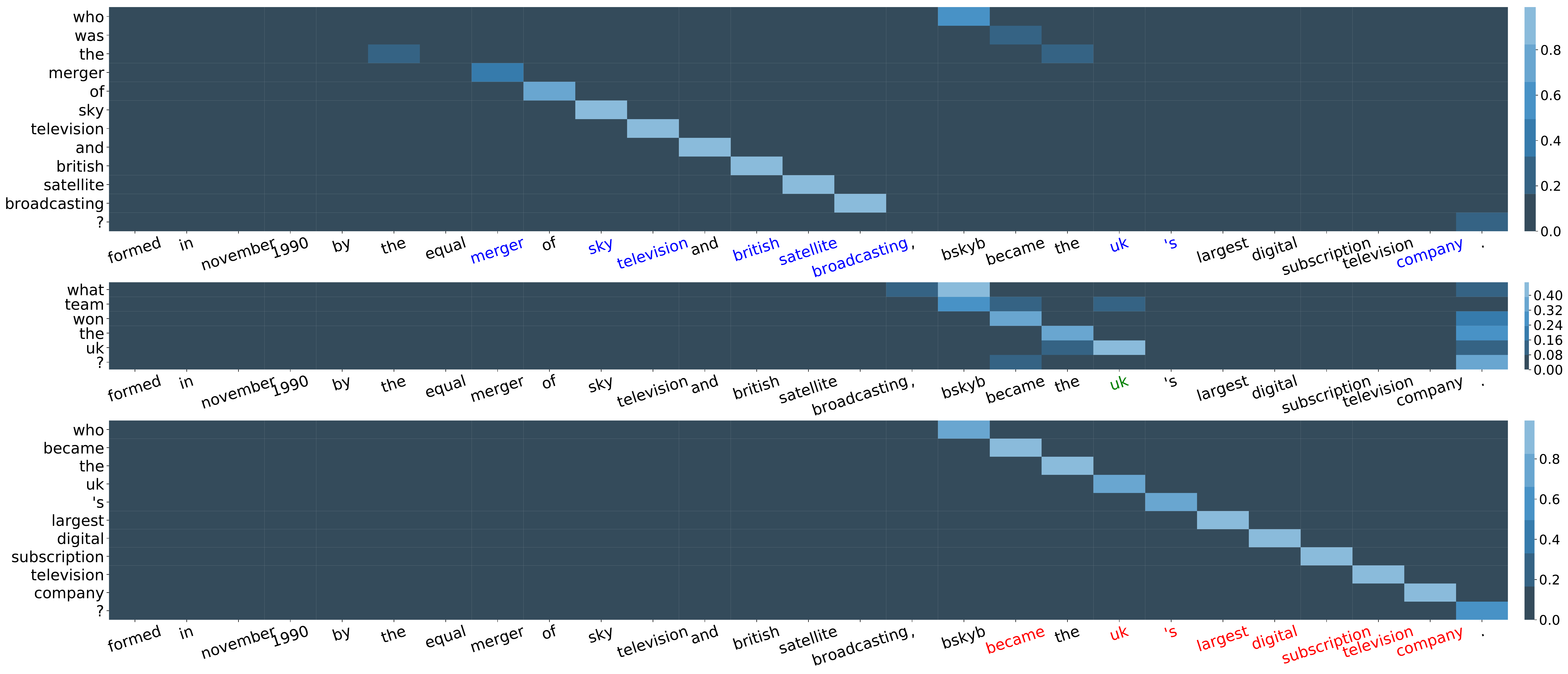}
\end{center} \caption{
    Attention heatmap of NQG++ decoder with three different focus by \Ours.
    This shows focus can guide generator to generate different sequences.
    Passage tokens are colored when corresponding focus is 1.
}
\label{fig:attention}
\end{figure*}

%% file: 05-results-t-human.tex
\begin{table}[!h]
    \begin{subtable}[c]{\columnwidth}
        \resizebox{\columnwidth}{!}{
        \begin{tabular}{l||ccc||ccc}
            \hline
             & \multicolumn{3}{c||}{Diversity (\%)} & \multicolumn{3}{c}{Accuracy (\%)} \\
            \hline
            \multicolumn{1}{c||}{Baselines} & Win & Lose & Tie & Win & Lose & Tie \\
            vs. 3-D. Beam & \textbf{49.7} & 31.3 & 19.0 & \textbf{43.9} & 36.9 & 19.2 \\
            vs. 3-T. Sampling & \textbf{46.7} & 35.1 & 18.2 & \textbf{45.3} & 36.1 & 18.6 \\
            vs. 3-M. Decoder & \textbf{47.6} & 32.5 & 19.9 & \textbf{41.8} & 36.0 & 22.2 \\
            \hline
        \end{tabular}
        }
        \caption{SQuAD question generation}
        \label{table:human_qg}
    \end{subtable}
    \begin{subtable}[c]{\columnwidth}
        \resizebox{\columnwidth}{!}{
        \begin{tabular}{l||ccc||ccc}
            \hline
             & \multicolumn{3}{c||}{Diversity (\%)} & \multicolumn{3}{c}{Accuracy (\%)} \\
            \hline
            \multicolumn{1}{c||}{Baselines} & Win & Lose & Tie & Win & Lose & Tie \\
            vs. 3-D. Beam & \textbf{50.4} & 40.9 & 8.7 & \textbf{46.2} & 38.5 & 15.3 \\
            vs. 3-T. Sampling & \textbf{48.7} & 42.0 & 9.3 & \textbf{50.3} & 41.2 & 8.5 \\
            vs. 3-M. Decoder & \textbf{49.7} & 39.6 & 10.7 & \textbf{46.5} & 37.5 & 16.0 \\
            \hline
        \end{tabular}
        }
        \caption{CNN-DM abstractive summarization}
        \label{table:human_sm}
    \end{subtable}
    \caption{Human evaluation results}
\end{table}

%% file: 05-results-t-single-expert.tex
\begin{table}[!h]
    \resizebox{\columnwidth}{!}{%
        \begin{tabular}{l||c|c|c}
            \hline
            \multicolumn{1}{c||}{Method} & R-1 & R-2 & R-L \\
            \hline
            PG \cite{See2017} & 39.53 & 17.28 & 36.38 \\
            Bottom-Up \cite{Gehrmann2018} & 41.22 & 18.68 & 38.34 \\
            DCA \cite{Celikyilmaz2018} & 41.69 & \textbf{19.47} & 37.92 \\
            \hline
            \Ours\ \& 10-Beam PG (Ours) & \textbf{41.72} & 18.74 & \textbf{38.79} \\
            \hline
    \end{tabular}
    }
    \caption{
        Comparison of single-expert selector with state-of-the-art abstractive summarization methods on CNN-DM.
        R stands for ROUGE \cite{Lin2004}
        }
    \label{table:single_expert}
\end{table}

%% file: 05-results-t-training-time.tex
\begin{table}[!h]
\begin{center}
    \resizebox{\columnwidth}{!}{%
        \begin{tabular}{l||c}
            \hline
            \multicolumn{1}{c||}{Method} & Training time (ms. / step) \\
            \hline
            PG & 641.2 \\ %
            3-M. Decoder & 1804.1 ($\times$ 2.81) \\ %
            5-M. Decoder & 2367.6 ($\times$ 4.37) \\ %
            \hline
            \Ours\ (Ours) & 692.1 ($\times$ 1.08)\\% # 823 / 2162.70 / 3125
            3-M. \Ours\ (Ours) & 740.8 ($\times$ 1.16)\\% # 800 / 2314.96/3125
            5-M. \Ours\ (Ours) & 747.6 ($\times$ 1.17) \\% # 815 / 2336.35/3125
            \hline
    \end{tabular}
    }
    \end{center}
    \caption{
        {\bf Training time:} Comparison of training time on CNN-DM. See \ref{exp_detail} for implementation details. 
        }
    \label{table:training_time}
\end{table}

%% file: 06-con.tex
We introduce a novel diverse sequence generation method via proposing a content selection module, \Ours.
Built upon mixture of experts and hard-EM training, \Ours\ identifies different key parts on source sequence to guide generator to output a diverse set of sequences.

\Ours\ is a generic plug-and-play module that can be added to an existing encoder-decoder model to enforce diversity with a negligible additional computational cost.
We empirically demonstrate that our method improves both accuracy and diversity and reduces training time significantly compared to baselines in question generation and abstractive summarization. Future work involves incorporating \Ours\ for other generation tasks such as diverse image captioning.